\documentclass[11pt]{article}

\usepackage[margin=1in]{geometry}
\usepackage{times}
\usepackage{microtype}
\usepackage{graphicx}
\usepackage{amsmath,amssymb}
\usepackage{booktabs}
\usepackage{float}
\usepackage{placeins}

\usepackage[colorlinks=true,citecolor=blue,linkcolor=blue,urlcolor=blue]{hyperref}

\title{Motion-Compensated Latent Semantic Canvases for Visual Situational Awareness on Edge}
\author{
	Igor Lodin \\
	\textit{AI and Computer Vision} \\
	\textit{Aimech Technologies Corp.}\\
	Kyiv, Ukraine\\
	igor.liodin@deepxhub.com
	\and
	Sergii Filatov \\
	\textit{AI Hardware and Computer Vision} \\
	\textit{Covijn Ltd.}\\
	London, United Kingdom\\
	sergii@deepxhub.com
	\and
	Vira Filatova \\
	\textit{Applied Artificial Intelligence} \\
	\textit{Covijn Ltd.}\\
	London, United Kingdom\\
	vira@deepxhub.com
	\and
    Dmytro Filatov \\
	\textit{Applied AI and Computer Vision} \\
	\textit{Aimech Technologies Corp.}\\
	San Francisco, USA\\
	dima@deepxhub.com\\
    (ORCID: 0000-0003-3836-5614)
}
\date{}

\begin{document}
\maketitle
\begin{abstract}
We propose \textbf{Motion-Compensated Latent Semantic Canvases (MCLSC)} for visual situational awareness on resource-constrained edge devices. The core idea is to maintain persistent semantic metadata in two latent canvases - a slowly accumulating \emph{static} layer and a rapidly updating \emph{dynamic} layer - defined in a baseline coordinate frame stabilized from the video stream. Expensive panoptic segmentation (Mask2Former) runs asynchronously and is \emph{motion-gated}: inference is triggered only when motion indicates new information, while stabilization/motion compensation preserves a consistent coordinate system for latent semantic memory. On prerecorded 480p clips, our prototype reduces segmentation calls by $>$30$\times$ and lowers mean end-to-end processing time by $>$20$\times$ compared to naive per-frame segmentation, while maintaining coherent static/dynamic semantic overlays.
\end{abstract}

\section{Introduction}
There is an increasing demand in edge devices providing AI-powered \emph{visual situational awareness}: persistent knowledge of what is in view (e.g., roads, sidewalks, buildings) and what is moving through the scene (e.g., people, vehicles). Modern panoptic segmentation models offer rich semantic structure \cite{cheng2021mask2former,kirillov2019panoptic}, but per-frame inference is often too expensive for real-time operation on embedded hardware - especially when additional compute is required for motion compensation and downstream control.

The second challenge is \emph{persistence under camera motion}. If semantic predictions are stored only in the current image frame, pixel coordinates change semantic meaning as the camera moves. It's crucial for an agent to remember ``where'' things were. In order to achieve this, semantics must be represented in a consistent coordinate system. For example, robotics systems integrate semantics into persistent maps \cite{mccormac2017semanticfusion,bescos2018dynaslam}, but they often assume heavier geometric estimation than is feasible or necessary for many edge use cases (wearables, dashcams, handheld AR).

This paper studies a practical question: \textbf{How can we maintain persistent semantic memory from video while applying heavy segmentation sparsely, maintaining compatibility with edge devices and satisfying real-time constraints?} We propose Motion-Compensated Latent Semantic Canvases (MCLSC), an architecture combining:
(i) baseline-anchored, feature-based stabilization (Lucas-Kanade tracking \cite{lucas1981iterative} with Shi-Tomasi features \cite{shitomasi1994good}) to estimate a stable transform into a baseline frame;
(ii) a 2$\times$ ``canvas'' warp to support motion compensation with reduced border artifacts; and
(iii) asynchronous, motion-gated panoptic segmentation using Mask2Former \cite{cheng2021mask2former}, whose outputs update two latent semantic layers: \emph{static} (accumulating) and \emph{dynamic} (replacing).

\textbf{Contributions.} (1) We formalize a lightweight representation for persistent semantics in video as dual latent canvases separated into static and dynamic classes. (2) We describe a motion-compensated update mechanism that keeps semantic coordinates stable under camera motion while permitting sparse inference. (3) We compare our approach to a naive baseline that performs segmentation on every frame, and report the resulting compute savings. Figure~\ref{fig:overview} summarizes the architecture.

\section{Related Work}
\paragraph{Panoptic and universal segmentation.}
Panoptic segmentation unifies semantic and instance segmentation into a single representation \cite{kirillov2019panoptic}. Prior work spans both bottom-up and set-prediction paradigms: Panoptic-DeepLab provides a simple and fast bottom-up baseline for panoptic segmentation \cite{cheng2020panopticdeeplab}, while MaskFormer and Mask2Former formulate segmentation as set prediction over masks \cite{cheng2021maskformer,cheng2021mask2former}. We use Mask2Former \cite{cheng2021mask2former} as an off-the-shelf panoptic model, but MCLSC is model-agnostic: any segmentation model producing labeled masks can update the canvases.

\paragraph{Semantic mapping under motion.}
Persistent semantic memory is a core goal in robotics. SemanticFusion integrates CNN semantics into a dense 3D map \cite{mccormac2017semanticfusion}, and DynaSLAM addresses dynamic scenes via segmentation-informed SLAM \cite{bescos2018dynaslam}. Our setting is lighter weight: rather than reconstructing 3D, we maintain 2D semantic canvases in a motion-compensated coordinate system derived from video stabilization, sufficient for situational awareness and as a substrate for downstream logic.

\paragraph{Video stabilization and motion compensation.}
Feature-based motion estimation is widely used for stabilization and mosaicing. We adopt a baseline-anchored approach based on good features to track \cite{shitomasi1994good} and pyramidal Lucas-Kanade optical flow \cite{lucas1981iterative}, with robust affine estimation via RANSAC. More sophisticated stabilization optimizes camera paths \cite{liu2013bundled}; our goal is a compute-efficient transform that supports persistent canvas coordinates.

\paragraph{Adaptive inference and conditional computation.}
Conditional computation reduces average inference cost by skipping computation or early exiting \cite{wang2018skipnet,ghodrati2021frameexit}. In video, temporal redundancy enables policies that avoid running expensive models on every frame. MCLSC applies a simple, interpretable policy: motion detection gates segmentation requests, and segmentation runs asynchronously so that stabilization and display remain responsive on edge devices.

\section{Problem Statement and Design Goals}
We consider an RGB video stream $\{I_t\}_{t=1}^T$ captured by a moving camera. The objective is to maintain persistent semantic context for situational awareness:
\begin{itemize}
    \item \textbf{Static context}: persistent environmental segments (e.g., road, sidewalk, sky, buildings).
    \item \textbf{Dynamic context}: transient objects (e.g., people, vehicles) that appear/disappear and move.
\end{itemize}

\paragraph{Problem.}
Naively running panoptic segmentation on every frame yields reliable masks but is too costly for real-time edge use and does not establish a stable coordinate frame. If segmentation is instead triggered only occasionally without motion compensation, the semantic masks become misaligned over time as the camera moves.

\paragraph{Design goals.}
\begin{enumerate}
    \item \textbf{Persistent coordinates:} write semantic masks into a stable baseline frame so that pixel coordinates have consistent meaning over time, enabling downstream logic to reason about \emph{where} something is even as the camera moves.
    \item \textbf{Dual-layer semantics:} separate static and dynamic semantics to support different temporal update rules (accumulate vs.\ replace), so long-lived structure can be retained while transient objects remain responsive to change.
    \item \textbf{Sparse segmentation:} avoid per-frame inference by triggering segmentation only when new information is likely (e.g., motion or scene change), reducing redundant computation while preserving semantic freshness when it matters.
    \item \textbf{Real-time compatibility:} enable bounded-latency operation via asynchronous inference and controlled frame dropping when compute falls behind, so the displayed state remains close to “now” instead of accumulating delay. 
    \item \textbf{Simplicity:} prefer interpretable components (feature tracking, motion thresholding) that are easy to tune on-device, debug in the field, and deploy across heterogeneous edge hardware without heavy dependencies.
\end{enumerate}

\section{Method}
\begin{figure}[H]
	\centering
	\includegraphics[width=0.75\linewidth]{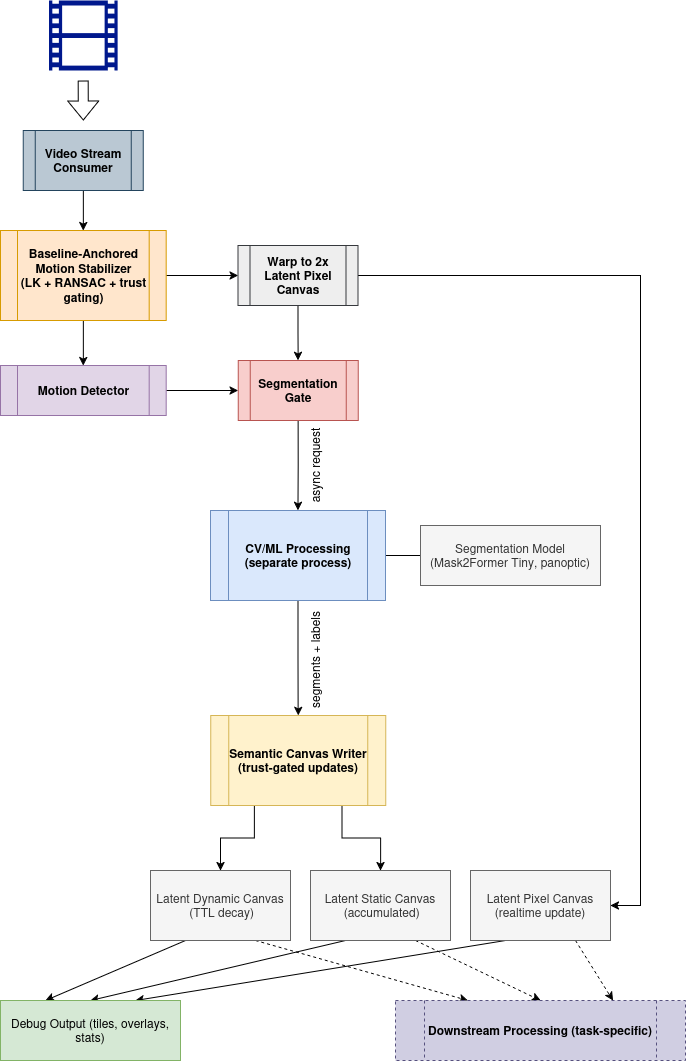}
	\caption{System Architecture of the Motion-Compensated Latent Semantic Canvases pipeline.}
	\label{fig:overview}
\end{figure}

Figure~\ref{fig:overview} provides an architectural overview of the proposed pipeline. We describe (i) baseline-anchored stabilization, (ii) canvas warping and latent semantic canvases, and (iii) motion-gated asynchronous segmentation.

\subsection{Baseline-Anchored Stabilization}
We estimate a 2D transform from the current frame to a baseline coordinate frame. Let $I_t$ be the current frame and $B$ the baseline frame (captured during initialization). We detect Shi-Tomasi corners \cite{shitomasi1994good} in $B$ (optionally using a dilated Canny edge mask to bias features toward strong gradients) and track them to $I_t$ using pyramidal Lucas-Kanade optical flow \cite{lucas1981iterative}. Given matched points, we robustly estimate an affine transform $A_t \in \mathbb{R}^{2\times3}$ via RANSAC and embed it into a 3$\times$3 matrix:
\[
T_t =
\begin{bmatrix}
A_t\\
0~0~1
\end{bmatrix},
\]
mapping coordinates in $I_t$ to coordinates in $B$.

\paragraph{Trust gating.}
Tracking can fail due to blur, occlusion, or low texture. We compute a confidence score from tracked feature count, inlier ratio, and median reprojection error, and only accept updates when the score exceeds a threshold. Otherwise we hold the last accepted transform. This prevents unstable warps from corrupting the semantic canvases.

\subsection{2$\times$ Canvas Warp}
Directly warping $I_t$ into $B$ produces border artifacts. We warp into a canvas that is 2$\times$ larger than the viewport. Let $C(\cdot)$ translate the viewport to the canvas center. We compute $M_t = C \, T_t$, warp the frame into the canvas, and crop the centered viewport. We also warp a binary validity mask to exclude padded regions from motion detection.

\subsection{Latent Static and Dynamic Semantic Canvases}
We maintain two label images in canvas coordinates:
\begin{itemize}
    \item \textbf{Static canvas} $S \in \mathbb{N}^{H\times W}$: accumulates static class labels; we write only where $S$ is empty.
    \item \textbf{Dynamic canvas} $D \in \mathbb{N}^{H\times W}$: represents dynamic labels in the current viewport; we clear and rewrite this region on each segmentation update.
\end{itemize}
Class labels are separated into \emph{static} and \emph{dynamic} via a curated taxonomy. Dynamic instances are tracked with IoU association and a frame TTL for pruning.

\subsection{Motion-Gated Asynchronous Segmentation}
Segmentation runs in a single worker thread. The main loop always performs stabilization, warping, motion detection, and visualization; segmentation results arrive asynchronously.

\paragraph{Motion gating.}
Let $V_t$ be the stabilized viewport and $m_t$ the valid mask. We compute a motion score from frame differencing:
\[
\Delta_t = |V_t - V_{t-1}|,\qquad
s_t = \operatorname{mean}(\Delta_t \odot m_t),
\]
and a motion area ratio $a_t$ from thresholded $\Delta_t$. We trigger segmentation when $(s_t > \tau_s)$ and $(a_t > \tau_a)$, subject to a minimum inter-call spacing.

\paragraph{Writing results.}
When the worker returns a panoptic segmentation $(P_t,\mathcal{S}_t)$, we resize $P_t$ (if needed) to the viewport size and update $S$ and $D$ in the canvas viewport region. Static segments fill empty static pixels; dynamic segments replace the dynamic region.

\paragraph{Real-time pacing and frame dropping.}
For bounded latency, we run the display loop at a fixed target frame rate and drop older frames from a bounded buffer when compute falls behind. After an initial prebuffer, this maintains approximately constant end-to-end latency while always displaying the most recent state.

\section{Discussion \& Limitations}
\paragraph{Why latent canvases?}
The dual-canvas representation provides a practical form of semantic memory: static context accumulates over time, while dynamic context reflects current movers. Motion compensation aligns updates into a persistent coordinate system without requiring full SLAM.

\paragraph{When does motion gating help most?}
Gating is most effective when temporal redundancy is high (mostly static environments with occasional movers). In highly dynamic scenes, segmentation frequency increases and compute savings shrink, but the approach still benefits from asynchronous execution and transform trust gating.

\paragraph{Accumulation errors.}
Static accumulation can ``lock in'' early mistakes. Confidence-weighted overwrites, slow decay, or periodic refresh segmentation are plausible improvements.

\paragraph{Geometric limits.}
Affine motion compensation is an approximation; parallax and non-planarity can cause misalignment in the stabilized canvas. Multi-plane models or lightweight SLAM could improve alignment at higher compute cost.

\paragraph{Implementation limits.}
Our prototype is Python-based (OpenCV + PyTorch/Transformers). Absolute timings vary by hardware; we emphasize relative comparisons between NAIVE and GATED under identical code paths.

\section{Experiments}
\subsection{Setup}
\begin{figure}[H]
	\centering
	\includegraphics[width=0.75\linewidth]{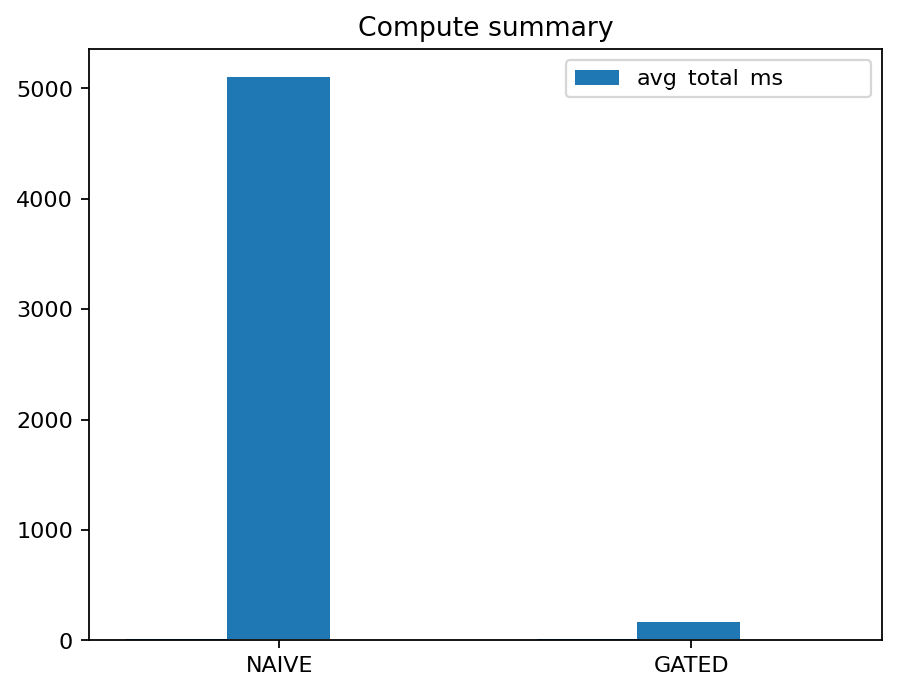}
	\caption{Aggregate runtime comparison between NAIVE and GATED (from evaluation logs).}
	\label{fig:bar_summary}
\end{figure}

\begin{figure}[H]
	\centering
	\includegraphics[width=0.75\linewidth]{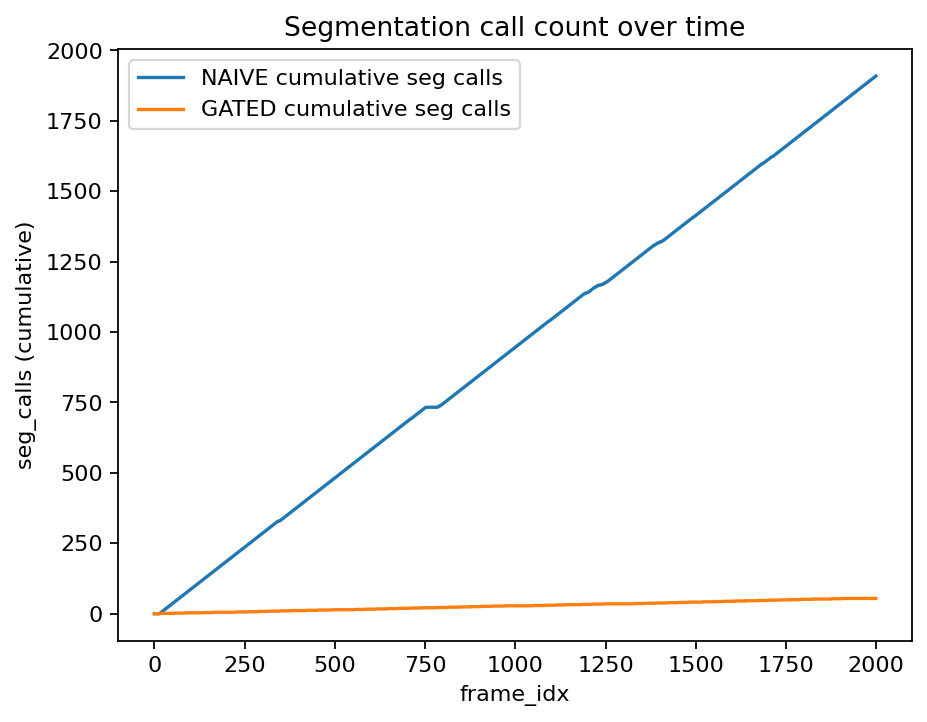}
	\caption{Segmentation submit events over time. NAIVE submits on (almost) every frame; GATED submits only when motion triggers it.}
	\label{fig:timeline_seg_calls}
\end{figure}

\begin{figure}[H]
	\centering
	\includegraphics[width=0.75\linewidth]{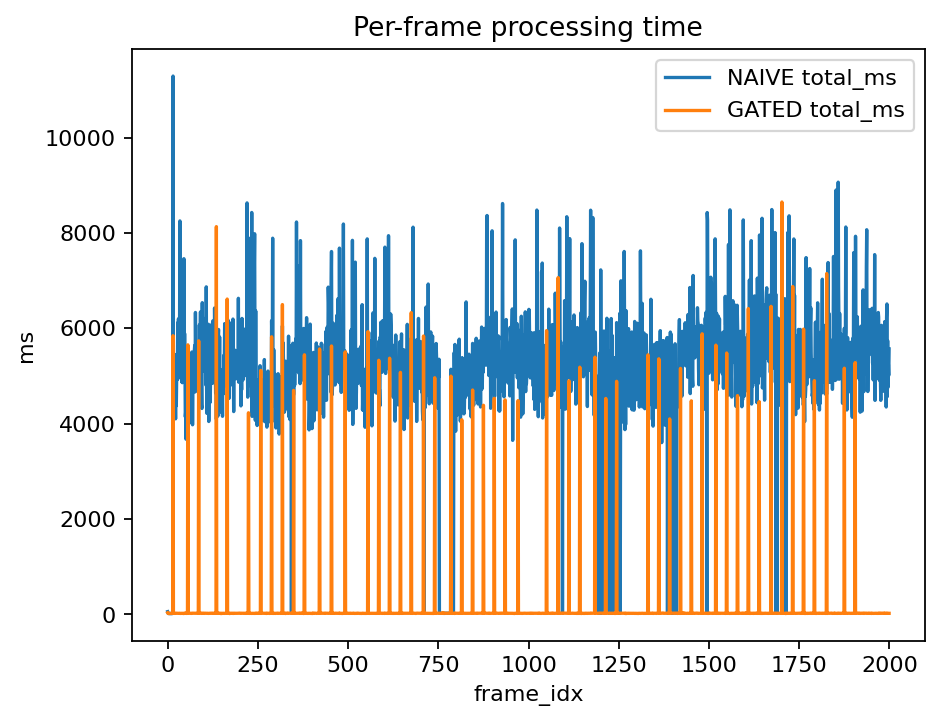}
	\caption{End-to-end processing time per frame. Spikes typically align with segmentation events.}
	\label{fig:timeline_total}
\end{figure}

We evaluate on prerecorded video clips processed at 854$\times$480. The pipeline runs stabilization and 2$\times$ canvas warping on every frame. The processing is CPU-based. We compare:
\begin{itemize}
    \item \textbf{NAIVE:} run panoptic segmentation on every frame, then update latent canvases.
    \item \textbf{GATED:} run panoptic segmentation only when motion triggers it (asynchronously), and update canvases when results arrive.
\end{itemize}
Both variants use Mask2Former-Swin-Tiny \cite{cheng2021mask2former}. We log per-frame timing for each stage, segmentation submit events, and derived throughput statistics. Figure~\ref{fig:bar_summary} and Table~\ref{tab:summary} summarize aggregates; Figures~\ref{fig:timeline_seg_calls} and \ref{fig:timeline_total} show time series.

\section{Metrics}
Let $t_i$ be the end-to-end processing time (ms) for frame $i$. Let $c_i \in \{0,1\}$ indicate whether segmentation was submitted for frame $i$. With $N$ frames:

\paragraph{Segmentation call count and rate.}
\[
N_{\text{seg}} = \sum_{i=1}^{N} c_i,\qquad
r_{\text{seg}} = \frac{N_{\text{seg}}}{N}.
\]

\paragraph{End-to-end processing statistics.}
\[
\mu_t = \frac{1}{N}\sum_{i=1}^{N} t_i,\qquad
p_q = \operatorname{percentile}_q(\{t_i\}_{i=1}^N),\ \ q\in\{50,95,99\}.
\]

\paragraph{Effective throughput.}
\[
\mathrm{FPS}_{\text{eff}} = \frac{1000}{\mu_t}.
\]

\paragraph{Savings ratios.}
For a metric $x$ (e.g., $\mu_t$ or $N_{\text{seg}}$), we report:
\[
\mathrm{speedup}(x) = \frac{x_{\text{NAIVE}}}{x_{\text{GATED}}}.
\]


\FloatBarrier 

\begin{table}[H]
	\centering
	\small
	\setlength{\tabcolsep}{5pt}
	\renewcommand{\arraystretch}{1.15}
	\begin{tabular}{lrrrrrrrr}
		\toprule
		Scenario & Frames & SegCalls & CallRate & Mean(ms) & Median & P95 & P99 & Eff.FPS \\
		\midrule
		NAIVE & 2000 & 1909 & 0.955 & 5100.1 & 5232.6 & 6630.5 & 8188.3 & 0.20 \\
		GATED & 2000 & 54 & 0.027 & 170.5 & 22.8 & 26.7 & 5630.1 & 5.86 \\
		\bottomrule
	\end{tabular}
	\caption{Runtime summary (ms) and segmentation call statistics from the provided evaluation logs.}
	\label{tab:summary}
\end{table}

\FloatBarrier 

\FloatBarrier 

\section{Results}
\subsection{Aggregate efficiency}
Table~\ref{tab:summary} reports runtime and segmentation call statistics. Compared to NAIVE, GATED reduces segmentation calls from nearly every frame to a small fraction of frames, and lowers mean end-to-end processing time correspondingly. Figure~\ref{fig:bar_summary} visualizes these aggregates.

\subsection{Temporal behavior}
Figure~\ref{fig:timeline_seg_calls} shows that NAIVE submits segmentation continuously, while GATED submits sparsely when motion triggers it. Figure~\ref{fig:timeline_total} shows the corresponding end-to-end processing time. NAIVE exhibits consistently high time per frame due to continuous inference. GATED has a low baseline with occasional spikes that align with segmentation events.

\subsection{Implications for edge operation}
MCLSC decouples semantic memory from per-frame inference: between segmentation events, the latent canvases provide usable semantic overlays in a stabilized coordinate system. Asynchronous execution and bounded buffering allow a fixed display rate; occasional inference spikes can be absorbed by dropping frames while preserving bounded latency.

\subsection{Qualitative visualization in debugging view}
\begin{figure}[H]
	\centering
	\includegraphics[width=\linewidth]{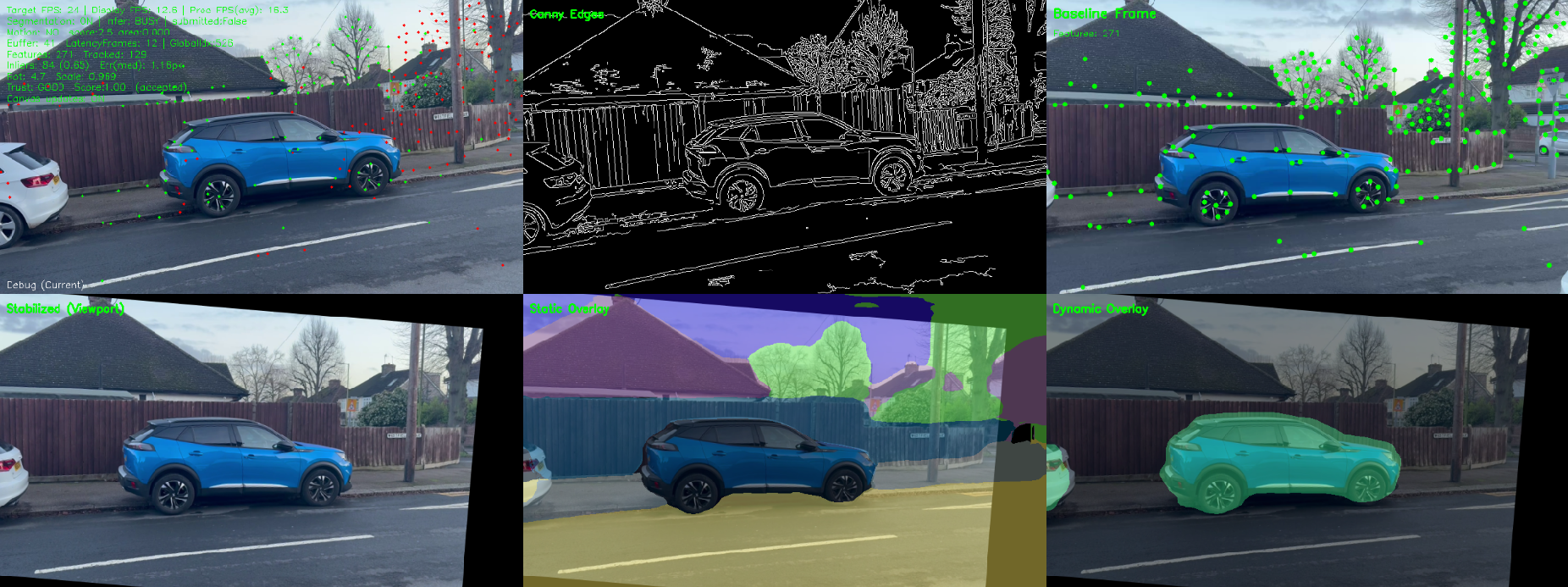}
	\caption{Runtime visualization (6 tiles): original input, edge map, baseline, stabilized viewport, static canvas overlay, and dynamic canvas overlay.}
	\label{fig:debug_tiles}
\end{figure}

Figure~\ref{fig:debug_tiles} shows the visual results that may be observed real time during video feed processing in OpenCV-based GUI.

\section{Ablations and Failure Modes}

\paragraph{Motion thresholds.}
The gating thresholds $(\tau_s,\tau_a)$ control how often segmentation runs, trading compute for semantic freshness. Lower thresholds trigger more segmentation updates, improving responsiveness to small changes but moving runtime toward the naive per-frame baseline. Higher thresholds reduce call rate and improve throughput, but can lead to stale masks (late detection of new objects, delayed removal of departed ones). Good settings depend on camera shake and scene dynamics.

\paragraph{Transform trust gating.}
Trust gating prevents stabilization failures from corrupting the canvases. If gating is too permissive, brief tracking failures (e.g., blur, occlusion, low texture) can produce incorrect transforms; writing semantics during these frames can stamp masks into wrong canvas locations and leave persistent artifacts. Holding the last accepted transform turns these periods into “no-update” windows, improving safety, but can temporarily delay semantic refresh until tracking recovers.

\paragraph{Static accumulation.}
The static canvas fills only previously empty regions to build long-term scene memory, but this can preserve early mistakes: a wrong static label may remain if that region is never rewritten. This is most visible during difficult initialization (occlusion, blur, or class confusion). Simple mitigations include periodic refresh, confidence-weighted overwrites, or marking low-confidence static regions for later correction.

\paragraph{Parallax and non-planar motion.}
We use a single affine warp for motion compensation, which is insufficient under strong parallax or large depth variation. Near objects and far background may require different warps, causing misalignment and ghosting in the canvases. This is a fundamental limitation of 2D stabilization without geometry; practical mitigations include limiting the operating regime, avoiding writes when parallax is high (via gating), or adding lightweight geometric cues when budget allows.

\section{Conclusions}
We introduced Motion-Compensated Latent Semantic Canvases (MCLSC), a practical architecture for maintaining persistent semantic situational awareness under camera motion and edge compute constraints. The core idea is to decouple \emph{semantic memory} from per-frame inference by (i) stabilizing incoming frames into a baseline coordinate system, (ii) storing semantics in latent canvas coordinates with explicit static/dynamic separation, and (iii) invoking panoptic segmentation only when motion indicates that new information is likely to appear. This design reduces reliance on continuous segmentation while preserving a coherent, persistent coordinate frame for semantic overlays, enabling mask persistence and incremental scene completion.

Our prototype demonstrates that large compute savings are achievable relative to naive per-frame segmentation, while retaining usable semantic continuity for typical handheld and lightly perturbed fixed-camera scenarios. At the same time, the method inherits limitations from image-plane motion models and model-dependent segmentation errors, especially under parallax, heavy blur, or early static misclassification. Future work includes confidence-weighted canvas writes, richer gating signals (e.g., feature-track statistics, optical-flow magnitude, or change detection on semantic logits), and integration with lightweight geometric mapping or layered motion to better handle depth variation. We also plan to explore lighter segmentation backbones and quantization strategies tailored to SBC-class deployment while preserving the static/dynamic canvas abstraction.

\bibliographystyle{plain}
\bibliography{refs}

\end{document}